\documentclass[a4paper]{article}

\usepackage{iclr2018_conference,times}
\usepackage{graphicx}
\usepackage{float}
\usepackage{multicol}
\usepackage{soul}
\usepackage{url}
\usepackage{amsmath}
\usepackage{amssymb}
\usepackage{booktabs}
\usepackage{adjustbox}
\title{Classification and Disease Localization in \\
       Histopathology Using Only Global Labels: \\
       A Weakly-Supervised Approach}
\author{Pierre Courtiol, Eric W. Tramel, Marc Sanselme, \& Gilles Wainrib \\
         Owkin, Inc. \\
         New York City, NY \\
         \texttt{\{firstname.lastname\}@owkin.com}}

\iclrfinalcopy

\begin{document}
\maketitle

\begin{abstract}
  Analysis of histopathology slides is a critical step for many diagnoses, and in particular in 
  oncology where it defines the gold standard. In the case of digital
  histopathological analysis, highly trained pathologists must review vast whole-slide-images of
  extreme digital resolution ($100,000^2$ pixels) across multiple zoom levels in order to locate
  abnormal regions of cells, or in some cases single cells, out of millions. The application of 
  deep learning to this problem is hampered not only by small sample sizes, as 
  typical datasets contain only a few hundred samples, 
  but also by the generation of ground-truth localized annotations for 
  training interpretable classification and segmentation models. We propose a method for disease
  localization in the context of weakly supervised learning, where only image-level labels are available during training. 
  Even without pixel-level annotations, we are able to
  demonstrate performance comparable with models trained with strong annotations on
  the Camelyon-16 lymph node metastases detection challenge. We accomplish this through
  the use of pre-trained deep convolutional networks, feature embedding, as well as learning via
  top instances and negative evidence, a multiple instance learning technique from
  the field of semantic segmentation and object detection.
\end{abstract}

\section{Introduction}
Histopathological image analysis (HIA) is a critical element of diagnosis in many areas of medicine, and
especially in oncology, where it defines the gold standard metric. 
Recent works
have sought to leverage modern developments in machine learning (ML) to aid pathologists in 
disease detection tasks, but the majority of these techniques require
localized annotation masks as training data. These annotations are 
even more costly to obtain than the original diagnosis, as pathologists must spend time to
assemble pixel-by-pixel segmentation maps of diseased tissue at extreme resolution, thus HIA 
datasets with annotations are very limited in size. Additionally, such localized annotations may
not be available when facing new problems in HIA, such as new disease subtybe classification, 
prognosis estimation, or drug response prediction. Thus, the critical question for HIA is:
can one design a learning architecture which achieves accurate classification with
no additional localized annotation? A successful technique would be able train algorithms to assist pathologists during analysis, and could also be used to identify previously unknown structures and regions
of interest.

Indeed, while histopathology is the gold standard diagnostic in oncology, it is 
extremely costly, requiring many hours of focus from pathologists to make a single diagnosis 
\citep{LST2016,Wea2010}. Additionally, as correct diagnosis for certain diseases requires pathologists to
identify a few cells out of millions, these tasks are akin to ``finding a needle in a haystack.''
Hard numbers on diagnostic error rates in histopathology are difficult to
obtain, being dependent upon the disease and tissue in question as well as
self-reporting by pathologists of diagnostic errors. 
However, as reported in the review of
\citet{SF2017}, false negatives in cancer diagnosis can lead not only to
catastrophic consequences for the patient, but also to incredible financial risk
to the pathologist. Any tool which can
aid pathologists to focus their attention and effort to the must suspect regions can help
reduce false-negatives and improve patient outcomes through more accurate diagnoses
\citep{DZA2017}. 
Medical researchers have looked to computer-aided diagnosis for decades, but the lack of 
computational resources and data have prevented wide-spread implementation and 
usage of such tools \citep{GBC2009}. Since the advent of automated digital WSI capture in the 1990s, researchers have sought approaches
for easing the pathologist's workload and improve patient outcomes through image processing
algorithms~\citep{GBC2009, LKB2017}. Rather than predicting final diagnosis, many of these procedures
focused instead on segmentation, either for cell-counting, or for the 
detection of suspect regions in the WSI. Historical methods have focused on the
use of hand-crafted texture or morphological \citep{DY2005} features
used in conjunction with unsupervised techniques such as K-means clustering or other dimensionality
reduction techniques prior to classification via k-Nearest Neighbor or a support vector machine. 

Over the past decade, fruitful developments in deep learning \citep{LBH2015}
have lead to an explosion of research into the automation of image processing tasks. While the application of such advanced ML techniques to image tasks has been successful for
many consumer applications, the adoption of such approaches within the field of medical 
imaging has been more gradual. However, these techniques demonstrate remarkable promise in 
the field of HIA. Specifically, in digital pathology with
whole-slide-imaging (WSI) \citep{YG2005,STM2016}, 
highly trained and skilled pathologists review digitally captured microscopy images from 
prepared and stained tissue samples in order to make diagnoses. 
Digital WSI are massive datasets, consisting of images captured at multiple
zoom levels. At the greatest magnification levels, a WSI may have a digital resolution upwards
of 100 thousand pixels in both dimensions. However, since localized annotations are very difficult 
to obtain, datasets may only contain WSI-level diagnosis labels, falling into the
category of weakly-supervised learning.

The use of DCNNs was first proposed for HIA in \citet{CGG2013}, where the
authors were able to train a model for mitosis detection in H\&E stained images. A similar
technique was applied for WSI for the detection of invasive ductal carcinoma in 
\citet{CBG2014}. These approaches demonstrated the usefulness of learned features as 
an effective replacement for hand-crafted image features. It is possible to train deep 
architectures from scratch for the classification of tile images \citep{WKG2016,HSK2016}. 
However, training such DCNN architectures can be extremely resource intensive. For this reason,
many recent approaches applying DCNNs to HIA make use of large pre-trained networks to act as rich 
feature extractors for tiles \citep{KMH2016, KCB2016,LST2016, XJW2017, SZC2017}. 
Such approaches have found success as aggregation of rich representations 
from pre-trained DCNNs has proven to be quite effective, 
even without from-scratch training on WSI tiles.

In this paper, we propose CHOWDER\footnote{%
Classification of HistOpathology with Weak supervision via Deep fEature aggRegation
}, an approach for the interpretable prediction of general localized 
diseases in WSI with only weak, whole-image disease labels and without any additional 
expert-produced localized annotations, i.e. per-pixel segmentation maps, 
of diseased areas within the WSI. 
To accomplish this, we modify an existing architecture from the field of multiple
instance learning and object region detection \citep{DTC2016} to 
WSI diagnosis prediction. By modifying the pre-trained DCNN model \citep{HZR2016}, 
introducing an additional set of fully-connected layers for context-aware classification from
tile instances, 
developing a random tile sampling scheme for efficient training over massive WSI, and
enforcing a strict set of regualrizations, we are able to demonstrate performance
equivalent to the best human pathologists \citep{BVD2017}.
Notably,
while the approach we propose makes use of a pre-trained DCNN as a feature extractor, 
the entire procedure 
is a true end-to-end classification technique, and therefore the transferred pre-trained
layers can be fine-tuned to the context of H\&E WSI. We demonstrate, 
using only whole-slide labels,
performance comparable to top-10 ranked methods trained with strong, pixel-level labels on 
the Camelyon-16 challenge dataset, while also producing disease segmentation that closely matches 
ground-truth annotations. We also present results for diagnosis prediction on WSI obtained from 
The Cancer Genome Atlas (TCGA), where strong annotations are not available and diseases
may not be strongly localized within the tissue sample.

\section{Learning Without Local Annotations}
While approaches using localized annotations 
have shown promise for HIA, they fail to address the cost associated
with the acquisition of hand-labeled datasets, as in each case these methods 
require access to pixel-level labels. As shown with ImageNet \citep{DDS2009}, access to data
drives innovation, however for HIA hand-labeled segmentation maps are costly
to produce, often subject to missed diseased areas, and cannot scale to the size of datasets required for truly effective deep learning.
Because of these considerations, HIA is uniquely suited to 
the \emph{weakly supervised learning} (WSL) setting. 

Here, we define the WSL task for HIA to be the identification of suspect regions of WSI when
the training data only contains image-wide labels of diagnoses made by expert pathologists. 
Since WSI are often digitally processed in small patches, or tiles, the aggregation of these
tiles into groups with a single label (e.g. ``healthy'', ``cancer present'') can be used 
within the framework of \emph{multiple instance learning} (MIL) 
\citep{DLL1997,Amo2013,XMF2014}. In MIL for binary classification, 
one often makes the standard multi-instance (SMI) assumption: a bag is classified as positive iff at
least \emph{one} instance (here, a tile) in the bag is labelled positive. The goal is to take the 
bag-level labels and learn a set of instance-level rules for the classification of single
instances. In the case of HIA, learning such rules provides the ability to infer localized regions
of abnormal cells within the large-scale WSI.

In the recent work of \citet{HSK2016} for WSI classification in the WSL setting, the 
authors propose an EM-based method to identify discriminative patches in high resolution 
images automatically during patch-level CNN training. They also introduced a decision level 
fusion method for HIA, which is more robust than max-pooling and can be thought of as a 
Count-based Multiple Instance (CMI) learning method with two-level learning.
While this approach was shown to be effective in the case of glioma classification and 
obtains the best result, it only slightly outperforms much simpler approaches presented in 
\citep{HSK2016}, but at much greater computational cost.

In the case of natural images, the WELDON and WILDCAT techniques of \citet{DTC2016} and
\citet{DMT2017}, respectively, demonstrated state-of-the-art performance for object
detection and localization for WSL with image-wide labels. In the case of
WELDON, the authors propose an end-to-end trainable CNN model based on MIL learning
with top instances \citep{LV2015} as well as negative evidence, relaxing the 
SMI assumption. Specifically, in the case of semantic segmentation, \citet{LV2015} 
argue that a target concept might not exist just at the subregion level, but that the
proportion of positive and negative samples in a bag have a larger effect in the determination of
label assignment. 
This argument also holds for the case of HIA, where pathologist diagnosis 
arises from a synthesis of observations across multiple resolution levels as well as the relative 
abundance of diseased cells. In Sec.~\ref{sec:chowder}, we will detail our proposed approach which makes
a number of improvements on the framework of \citet{DTC2016}, adapting it to the context of 
large-scale WSI for HIA.

\subsection{WSI Pre-Processing}

\paragraph{Tissue Detection.}
As seen in Fig. \ref{fig:baseline-schema}, large regions of a WSI may contain no tissue at all, and
are therefore not useful for training and inference. To extract only tiles with content relevant to 
the task, we use the same approach as \citet{WKG2016}, namely, Otsu's method \citep{Ots1979} applied to the 
hue and saturation channels of the image after transformation into the HSV color space to produce
two masks which are then combined to produce the final tissue segmentation.
Subsequently, only tiles within the foreground segmentation are extracted for training and inference.

\paragraph{Color Normalization.}
According to \citet{CGB2017}, stain normalization is an important step in HIA 
since the result of the H\&E staining procedure can vary greatly between any two slides. 
We utilize a simple histogram equalization algorithm consisting of left-shifting RGB channels and 
subsequently rescaling them to $[0, 255]$, as proposed in \citet{Whitebal}. 
In this work, we place a particular emphasis on the tile aggregation method 
rather than color normalization, so we did not make use of more advanced color normalization algorithms,
such as \citet{KRT2014}.

\paragraph{Tiling.}
The tiling step is necessary in histopathology analysis. Indeed, due to the large size of the WSI, 
it is computationally intractable to process the slide in its entirety. For example, on the highest 
resolution zoom level, denoted as \emph{scale 0}, for a fixed grid of non-overlapping tiles, a WSI may possess more 
than 200,000 tiles of $224\times 224$ pixels. Because of the computational burden associated with 
processing the set of all possible tiles, we instead turn to a uniform random sampling from the space 
of possible tiles. Additionally, due to the large scale nature of WSI datasets, the computational 
burden associated with sampling potentially overlapping tiles from arbitrary locations is a prohibitive 
cost for batch construction during training. 

Instead, we propose that all tiles from the non-overlapping grid should be processed and stored to disk 
prior to training. As the tissue structure does not exhibit any strong 
periodicity, we find that sampling tiles along a fixed grid without overlapping provides a  
reasonably representative sampling while maximizing the total sampled area.

Given a target scale $\ell \in \{0, 1, \dots, L\}$, we denote the number of possible tiles in
WSI indexed by $i \in \{1, 2, \dots, N \}$ as $M^{\rm T}_{i, \ell}$. The number of tiles sampled for 
training or inference is denoted by $M^{\rm S}_{i, \ell}$ and is chosen according to
\begin{equation}
M^{\rm S}_{i,\ell} = 
\min\left(M^{\rm T}_{i,\ell}~,~\max\left(M^{\rm T}_{\rm min},~\frac{1}{2}\cdot \bar{M}^{\rm T}_{\ell}\right) \right),    
\end{equation}
where $\bar{M}^{\rm T}_{\ell} = \frac{1}{N}\sum_i M^{\rm T}_{i,\ell}$ is the empirical average of the
number of tiles at scale $\ell$ over the entire set of training data. 

\paragraph{Feature Extraction.}
We make use of the ResNet-50 \citep{HZR2016} architecture trained on the ImageNet natural image dataset. 
In empirical comparisons between VGG or Inception architectures, we have found that the ResNet 
architecture provides features more well suited for HIA. Additionally, the ResNet architecture
is provided at a variety of depths (ResNet-101, ResNet-152). However, we found that ResNet-50
provides the best balance between the computational burden of forward inference and richness of
representation for HIA.

In our approach, for every tile we use the values of the ResNet-50 pre-output layer, a set of $P = 2048$ 
floating point values, as the feature vector for the tile. Since the fixed input resolution for 
ResNet-50 is $224\times 224$ pixels, we set the resolution for the tiles extracted from the WSI to
the same pixel resolution at every scale $\ell$.

\subsection{Baseline method}
Given a WSI, extracting tile-level features produces a bag of feature vectors which one attempts to
use for classification against the known image-wide label. The dimension of these local descriptors
is $M^{\rm S} \times P$, where $P$ is the number of features output
from the pre-trained image DCNN and $M^{\rm S}$ is the number of sampled tiles.
Approaches such as Bag-of-visual-words (BoVW) or VLAD \citep{JDS2010} could be chosen as a baseline
aggregation method to generate a single image-wide descriptor of size $P\times 1$, but would require
a huge computational power given the dimensionality of the input.
Instead, we will try two common approaches for the aggregation of local features, specifically, the 
\texttt{MaxPool} and \texttt{MeanPool} and subsequently apply a classifier on the aggregated features.
After applying these pooling methods over the axis of tile indices, one obtains a single feature 
descriptor for the whole image. Other pooling approaches have been used in the context of HIA, 
including Fisher vector encodings \citep{SZC2017} and $p-$norm pooling \citep{XJW2017}.
However, as the reported effect of these aggregations is quite small, we don't consider these 
approaches when constructing our baseline approach. 

\begin{figure}[t]
    \centering
    \includegraphics[width=10cm,height=20cm,keepaspectratio]{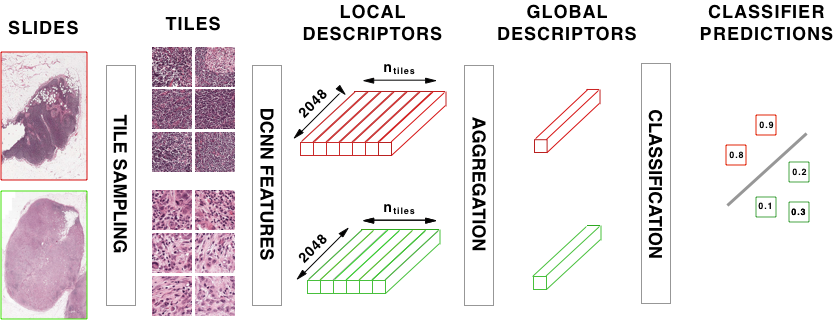}
    \caption{\label{fig:baseline-schema}%
             Description of the BASELINE approach for WSI classification via aggregation of tile-level
             features into global slide descriptors.}
\end{figure}

After aggregation, a classifier can be trained to produce the desired diagnosis labels given the
global WSI aggregated descriptor. For our baseline method, we use a logistic regression for this
final prediction layer of the model. We present a description of the baseline approach in 
Fig. \ref{fig:baseline-schema}.

\subsection{CHOWDER Method}
\label{sec:chowder}
\begin{figure}[t]
    \centering
    \includegraphics[width=\textwidth,keepaspectratio]{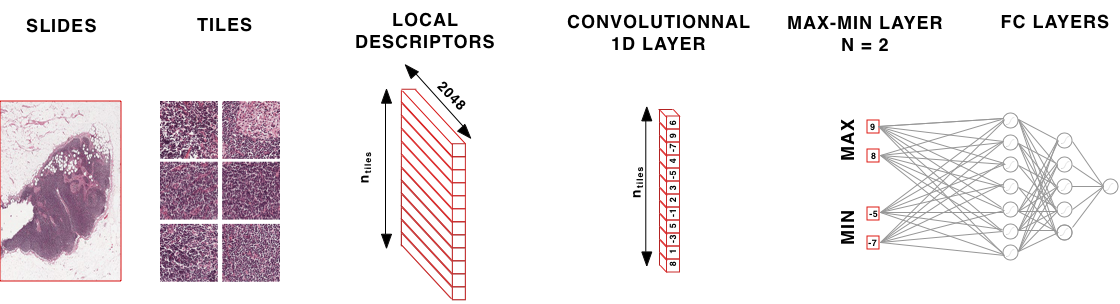}
    \caption{\label{fig:chowder-schema}%
             Description of the CHOWDER architecture (for $R=2$) for WSI classification via
             MLP on operating on top positive and negative instances shown for a single
             sample mini-batch sample. }
\end{figure}

In experimentation, we observe that the baseline approach of the previous section works well for 
\emph{diffuse} disease, which is evidenced in the results of Table \ref{tab:full-results} for
\texttt{TCGA-Lung}. Here, \emph{diffuse} implies that the number of disease-containing tiles,
pertinent to the diagnosis label, are roughly proportional to the number of tiles containing 
healthy tissue. However, if one applies the same approach to different WSI datasets, such as 
\texttt{Camelyon-16}, the performance significantly degrades. In the case of \texttt{Camelyon-16},
the diseased regions of most of the slides are highly localized, restricted to a very small area 
within the WSI. When presented with such imbalanced bags, simple aggregation approaches for global
slide descriptors will overwhelm the features of the disease-containing tiles.

Instead, we propose an adaptation and improvement of the WELDON method \citep{DTC2016}
designed for histopathology images analysis. As in their approach, rather than creating a 
global slide descriptor by aggregating all tile features, instead a MIL approach is used that
combines both top-instances as well as negative evidence. A visual description of approach is
given in Fig. \ref{fig:chowder-schema}.

\paragraph{Feature Embedding.}
First, a set of one-dimensional embeddings for the $P = 2048$ ResNet-50 features are calcualted via
$J$ one-dimensional convolutional layers strided across the tile index axis. 
For tile $t$ with features $\mathbf{k}_t$, the embedding according to kernel $j$ is calculated 
as $e_{j,t} = \left<  \mathbf{w}_j, \mathbf{k}_t\right>$. Notably, the kernels $\mathbf{w}_j$ have dimensionality 
$P$. This one-dimensional convolution is, in essence, a
shortcut for enforcing a fully-connected layer with tied weights across tiles, i.e. the same embedding
for every tile \citep{DTC2016}. 
In our experiments, we found that the use of a single embedding, $J=1$, is an appropriate choice
for WSI datasets when the number of available slides is small ($<1000$). In this case, choosing
$J>1$ will decrease training error, but will \emph{increase} generalization error. 
Avoiding overtraining and ensuring model generality remains a major challenge for the application
of WSL to WSI datasets.

\paragraph{Top Instances and Negative Evidence.}
After feature embedding, we now have a $M^{\rm S}_{\ell, i} \times 1$ vector of
local tile-level (\emph{instance}) descriptors. As in \citep{DTC2016}, these instance descriptors 
are sorted by value. Of these sorted embedding values, only the top and bottom $R$ entries are 
retained, resulting in a tensor of $2R \times 1$ entries to use for diagnosis classification. This can
be easily accomplished through a \texttt{MinMax} layer on the output of the one-dimensional convolution
layer.
The purpose of this layer is to take not only the top instances region but also the negative
evidences, that is the region which best support the absence of the class. During training, the 
back-propagation runs only through the selected tiles, positive and negative evidences. 
When applied to WSI, the \texttt{MinMax} serves as a powerful tile selection procedure.

\paragraph{Multi-layer Perceptron (MLP) Classifier.}
In the WELDON architecture, the last layer consists of a sum applied over the $2R\times 1$ output
from the \texttt{MinMax} layer. However, we find that this approach can be improved for 
WSI classification. We investigate the possibility of a richer interactions between the
top and bottom instances by instead using an MLP as the final classifier. In our implementation of CHOWDER, we use
an MLP with two fully connected layers of 200 and 100 neurons with sigmoid activations.

\section{Experimental Results}
\subsection{Training Details}
First, for pre-processing, we fix a single tile scale for all methods and datasets. 
We chose a fixed zoom level of 0.5 $\mu$m/pixel, which corresponds to $\ell=0$ for slides scanned
at 20x magnification, or $\ell=1$ slides scanned at 40x magnification. Next, since WSI datasets often only contain
a few hundred images, far from the millions images of ImageNet dataset, strong regularization 
required prevent over-fitting. 
We applied $\ell_2$-regularization of 0.5 on the convolutional feature embedding layer and 
dropout on the MLP with a rate of 0.5. However, these values may not be the global optimal, as we 
did not apply any hyper-parameter optimization to tune these values. To optimize the model parameters,
we use Adam \citep{KB2014} to minimize the binary cross-entropy loss over 30 epochs with a mini-batch size 
of 10 and with learning rate of 0.001.

To reduce variance and prevent over-fitting, we trained an ensemble of $E$ CHOWDER 
networks which only differ by their initial weights. The average of the predictions made 
by these $E$ networks establishes the final prediction. Although we set $E = 10$ for the results 
presented in Table~s\ref{tab:full-results}, we used a larger ensemble of $E = 50$ with $R=5$ 
to obtain the best possible model when comparing our method to those 
reported in the \texttt{Camelyon-16} leaderboards.
We also use an ensemble of $E = 10$ when reporting the results for WELDON.
As the training of one epoch requires about 30 seconds on our available hardware, 
the total training time for the ensemble took just over twelve hours. 
While the ResNet-50 features were extracted using a GPU for efficient feed-forward calculations, 
the CHOWDER network is trained on CPU in order to take advantage of larger system RAM sizes, 
compared to on-board GPU RAM. This allows us to store all the training tiles in memory to provide
faster training compared to a GPU due to reduced transfer overhead. 

\subsection{TCGA}
The public Cancer Genome Atlas (TCGA) provides approximately 11,000 tissue slides images of 
cancers of various organs\footnote{%
\url{https://portal.gdc.cancer.gov/legacy-archive}
}. For our first experiment, we selected 707 lung cancer WSIs (\texttt{TCGA-Lung}), which were
downloaded in March 2017. Subsequently, a set of new lung slides have been added to TCGA, increasing
the count of lung slides to 1,009. Along with the slides themselves, TCGA also provides
labels representing the type of cancer present in each WSI. However,
no local segmentation annotations of cancerous tissue regions are provided. 
The pre-processing step extracts 1,411,043 tiles and their corresponding representations from ResNet-50.
The task of these experiments is then to predict which type of cancer is contained in each WSI: 
adenocarcinoma or squamous cell carcinoma. We evaluate the quality of the classification according to the
area under the curve (AUC) of the receiver operating characteristic (ROC) curve generated using 
the raw output predictions. 

As expected in the case of diffuse disease, the advantage provided by CHOWDER is slight as compared to
the \texttt{MeanPool} baseline, as evidenced in Table~\ref{tab:full-results}.
Additionally, as the full aggregation techniques work quite well in this
setting, the value of $R$ does not seem to have a strong effect on the performance of CHOWDER as it 
increases to $R = 100$. In this setting of highly homogenous tissue content, we can expect that global
aggregate descriptors are able to effectively separate the two classes of carcinoma. 

\subsection{Camelyon-16}
For our second experiment,
we use the \texttt{Camelyon-16} challenge dataset\footnote{%
\url{https://camelyon16.grand-challenge.org}
}, 
which consists of 400 WSIs taken from sentinel lymph nodes, 
which are either healthy or exhibit metastases of some form. In addition to the WSIs 
themselves, as well as their labeling (\texttt{healthy}, \texttt{contains-metastases}), 
a segmentation mask is provided for each WSI which 
represents an expert analysis on the location of metastases within the WSI. Human 
labeling of sentinel lymph node slides is known to be quite tedious, as noted in \citet{LST2016,Wea2010}.
Teams participating in the challenge had access to, and utilized, the ground-truth masks when training
their diagnosis prediction and tumor localization models. For our approach, we set aside the masks of 
metastasis locations and utilize only diagnosis labels.
Furthermore, many participating teams developed a post-processing step, extracting handcrafted 
features from predicted metastasis maps to improve their segmentation. No post-processing is 
performed for the presented CHOWDER results,
the score is computed directly from the raw output of the CHOWDER model.

The \texttt{Camelyon-16} dataset is evaluated on two different axes. First, the accuracy of the 
predicted label for each WSI in the test set is evaluated according to AUC.
Second, the accuracy of metastasis localization is evaluated by comparing model outputs to the
ground-truth expert annotations of metastasis location. This segmentation accuracy is measured according
to the free ROC metric (FROC), which is the curve of metastasis detection sensitivity to the 
average number of also positives. As in the Camelyon challenge, we evaluate
the FROC metric as the average detection sensitivity at the average false positive rates 
0.25, 0.5, 1, 2, 4, and 8.  

\paragraph{Competition Split Bias.} 
We also conduct a set of experiments on \texttt{Camelyon-16} using random train-test cross-validation
(CV) splits, respecting
the same training set size as in the original competition split. We note distinct difference in AUC 
between the competition split and those obtained via random folds. This discrepancy is especially
distinct for the \texttt{MeanPool} baseline, as reported in Table~\ref{tab:full-results}. We therefore
note a distinct discrepancy in the data distribution between the competition test and training splits. 
Notably, using the \texttt{MeanPool} baseline architecture, we found that 
the competition train-test split can be predicted with an AUC of 0.75, however one only obtains an
AUC of 0.55 when using random splits. Because this distribution mismatch in the competition split could
produce misleading interpretations, we report 3-fold average CV results along with the results obtained
on the competition split.

\paragraph{Classification Performance.}
In Table~\ref{tab:full-results}, we see the classification performance of our proposed CHOWDER 
method, for $E = 10$, as compared to both the baseline aggregation techniques, as well as the WELDON 
approach. In the
case of WELDON, the final MLP is not used and instead a summing is applied to the \texttt{MinMax} layer.
The value of $R$ retains the same meaning in both cases: the number of both high and low scoring tiles
to pass on to the classification layers. We test a range of values $R$ for both WELDON and CHOWDER.
We find that over all values of $R$, CHOWDER provides a significant advantage over both the baseline
aggregation techniques as well as WELDON. We also note that the optimal performance can be obtained 
without using a large number of discriminative tiles, i.e. $R = 5$. 

\begin{figure*}
    \centering
    \includegraphics[width=0.4\textwidth, keepaspectratio]{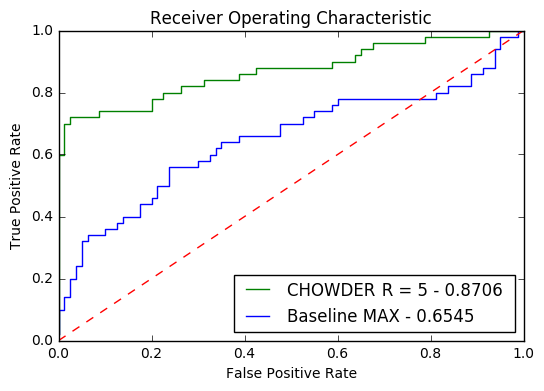}
    \quad
    \includegraphics[width=0.4\textwidth, keepaspectratio]{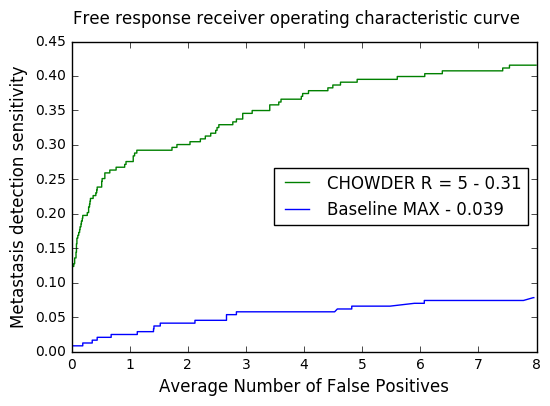}
    \caption{\label{fig:camelyon-curves}%
    Performance curves for \texttt{Camelyon-16} dataset 
    for both classification and segmentation tasks for the different tested approaches. 
    \textbf{Left:} ROC curves for the classification task. 
    \textbf{Right:} FROC curves for lesion detection task.
    }
\end{figure*}

Of the published results on \texttt{Camelyon-16}, the winning approach of \citet{WKG2016} 
acheived an AUC of 0.9935. With CHOWDER we are able to attain an AUC of 0.8706,
\emph{but without using any of the ground-truth disease segmentation maps}. This is a remarkable 
result, as \citet{WKG2016} required tile-level disease labels derived from
expert-provided annotations in order to train a full 27-layer GoogLeNet \citep{SLJ2015} 
architecture for tile-level tumor prediction. We also note that CHOWDER's performance on this task 
roughly is equivalent to the best-performing human pathologist, an AUC of 0.884 as reported
by \citet{BVD2017}, and better-than-average with respsect to human pathologist performance, 
an AUC of 0.810. Notably, this human-level performance is achieved \emph{without} human 
assistance during training, beyond the diagnosis labels themselves. 
We present the ROC curve for this result in Fig.~\ref{fig:camelyon-curves}. 

\paragraph{Localization Performance.}
Obtaining high performance in terms of whole slide classification is well and good, but it is 
not worth much without an interpretable result which can be used by pathologists to aid their 
diagnosis. For example, the \texttt{MeanPool} baseline aggregation approach provides no information
during inference from which one could derive tumor locations in the WSI: all locality information is
lost with the aggregation. With \texttt{MaxPool}, one at least retains some information via the 
tile locations which provide each maximum aggregate feature. 

\begin{figure*}[t]
    \centering
    \begin{minipage}[c]{0.34\textwidth}
        \fbox{\includegraphics[width=\textwidth, trim={2.5cm 13.75cm 5cm 27.25cm}, clip]{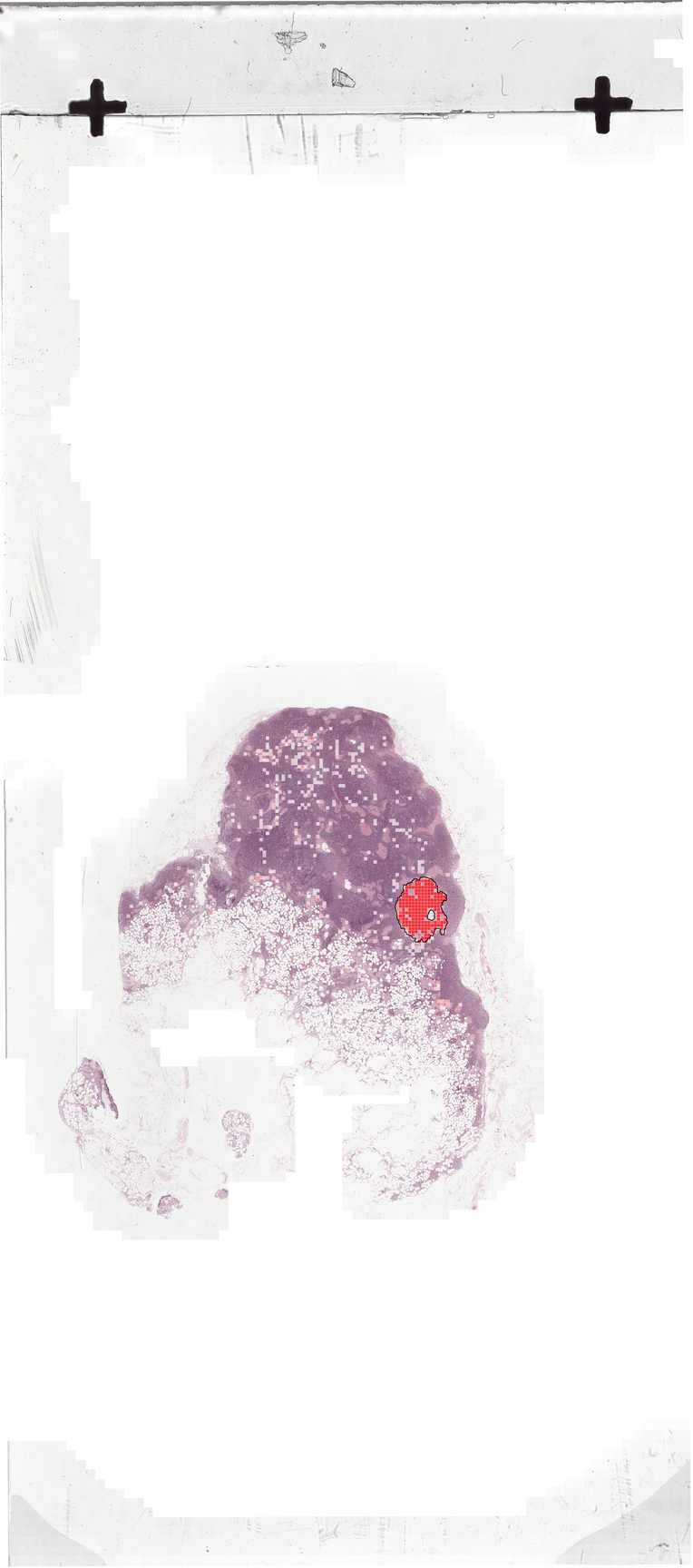}}
    \end{minipage}        
    \quad
    \begin{minipage}[c]{0.35\textwidth}
        \fbox{\includegraphics[width=\textwidth]{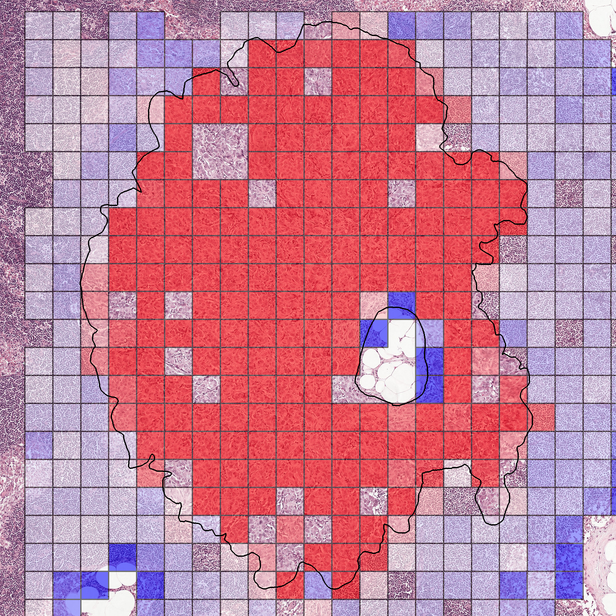}}
    \end{minipage}
    
    \caption{\label{fig:heatmap-27}%
            Visualization of metastasis detection on test image 27 of the \texttt{Camelyon-16} dataset 
            using our proposed approach. 
            \textbf{Left:} Full WSI at zoom level 6 with ground truth
            annotation of metastases shown via black border. Tiles with positive feature embeddings
            are colored from white to red according to their magnitude, with red 
            representing the largest magnitude.
            \textbf{Right:} Detail of metastases at zoom level 2 overlaid with classification 
            output of our proposed approach. Here, the output of all tested tiles are shown and
            colored according to their value, from blue to white to red, with blue representing
            the most negative values, and red the most positive.
            Tiles without color were not included when randomly selecting tiles for inference.}
\end{figure*}

For CHOWDER, we propose the use of the full set of outputs from the convolutional feature embedding layer.
These are then sorted and thresholded according to value $\tau$ such that tiles with an embedded value
larger than $\tau$ are classified as diseased and those with lower values are classified as healthy.
We show an example of disease localization produced by CHOWDER in Fig. \ref{fig:heatmap-27}. Here,
we see that CHOWDER is able to very accurately localize the tumorous region in the WSI 
even though it has only been trained using global slide-wide labels and without any local annotations. 
While some potential false detections occur outside of the tumor region, we see that the strongest
response occurs within the tumor region itself, and follows the border regions nicely. We present
further localization results in Appendix \ref{sec:further-results}.

Comparing to the published \texttt{Camelyon-16} results, the approach of \cite{WKG2016} achieved
a winning FROC of 0.8074 with strong annotations. In the weak-leaning setting, CHOWDER 
achieves a FROC of 0.3103. However, we note that these localization results are preliminary, as we
do not conduct either a full sampling of all tiles in the WSI, nor do we apply post-processing
of overlapping tiles to obtain full-resolution localization maps. While approaches are certainly
possible using tile-instanace feature embeddings obtained from the trained CHOWDER architecture, 
we note that further innovations in post-processing are outside the scope of this work.
The FROC curve is given in Fig. \ref{fig:camelyon-curves}.

\begin{table}[t!]
\centering
\begin{minipage}[c]{0.45\textwidth}\small
    \centering
    \begin{tabular}{lccc}
    \toprule
    &  & \multicolumn{2}{c}{\bf AUC} \\
    \cmidrule{3-4}
    \multicolumn{1}{l}{\bf Method}& & {\it CV} & {\it Competition} \\
    \midrule

    \multicolumn{1}{l}{\it BASELINE} &  & & \\
    \cmidrule{1-1}
    {\texttt{MaxPool}}  & & 0.749 & 0.655 \\
    {\texttt{MeanPool}} &  & 0.802 & 0.530\vspace{2ex}\\

    \multicolumn{1}{l}{\it WELDON} &  & & \\
    \cmidrule{1-1}
    {$R = 1$}  & & 0.782 & 0.765 \\
    {$R = 10$} &  & 0.832 & 0.670 \\
    {$R = 100$}  & & 0.809 & 0.600 \\
    {$R = 300$}  & & 0.761 & 0.573 \vspace{2ex}\\

    \multicolumn{1}{l}{\it CHOWDER} &  & & \\
    \cmidrule{1-1}

    {$R = 1$} &  &  0.809 & 0.821 \\
    {$R = 5$}  & & \textbf{0.903} & \textbf{0.858} \\
    {$R = 10$}  & & 0.900 & 0.843 \\
    {$R = 100$} &  & 0.870 & 0.775 \\
    {$R = 300$} &  & 0.837 & 0.652 \\

    \bottomrule
    \end{tabular}
\end{minipage}
\begin{minipage}[c]{0.45\textwidth}\small
    \centering
    \begin{tabular}{lcc}
        \toprule
        {\bf Method} &  & {\bf AUC} \\
        \midrule
        
        \multicolumn{1}{l}{\it BASELINE} &  & \\
        \cmidrule{1-1}
        {\texttt{MaxPool}}& &  0.860 \\
        {\texttt{MeanPool}} &  & 0.903\vspace{2ex}\\
        
        \multicolumn{1}{l}{\it CHOWDER} & & \\
        \cmidrule{1-1}
        {$R = 1$} &  &   0.900 \\
        {$R = 10$}  & &  \textbf{0.915} \\
        {$R = 100$} &  & 0.909 \\
        
        \bottomrule
        \end{tabular}
\end{minipage}
\caption{\label{tab:full-results}
         Classification (AUC) results for the \texttt{Camelyon-16} (\textbf{left}) and 
         \texttt{TCGA-Lung} (\textbf{right}) datasets
         for CHOWDER, WELDON, and the baseline approach. 
         For \texttt{Camelyon-16}, we present two scores, one for the fixed competition 
         test split of 130 WSIs, 
         and one for a cross-validated average over 3 folds (\emph{CV}) on the 270 training WSIs.
         For \texttt{TCGA-Lung}, we present scores as a cross-validated average
         over 5 folds.}
\end{table}

\section{Discussion}
We have shown that using state-of-the-art techniques from MIL in computer vision, such as the 
top instance and negative evidence approach of \citep{DTC2016}, one can construct an effective 
technique for diagnosis prediction \emph{and} disease location for WSI in histopathology without
the need for expensive localized annotations produced by expert pathologists. By removing this 
requirement, we hope to accelerate the production of computer-assistance tools for pathologists to
greatly improve the turn-around time in pathology labs and help surgeons and oncologists make rapid
and effective patient care decisions. This also opens the way to tackle problems where 
expert pathologists may not know precisely where relevant tissue is located within the slide image, 
for instance for prognosis estimation or prediction of drug response tasks. The ability of our 
approach to discover associated regions of interest without prior localized annotations hence 
appears as a novel discovery approach for the field of pathology. Moreover, using the suggested 
localization from CHOWDER, one may considerably speed up the process of obtaining ground-truth 
localized annotations.

A number of improvements can be made in the CHOWDER method, 
especially in the production of disease localization maps. As presented, we use the raw values from
convolutional embedding layer, which means that the resolution of the produced disease localization 
map is fixed to that of the sampled tiles. However, one could also sample overlapping tiles and then 
use a data fusion technique to generate a final localization map. Additionally, as a variety of 
annotations may be available, CHOWDER could be extended to the case of heterogenous annotation,
e.g. some slides with expert-produced localized annotations and those with only whole-slide
annotations.

\clearpage
\small
\bibliography{references}
\bibliographystyle{iclr2018_conference}

\clearpage
\normalsize
\appendix
\section{Further Results}
\label{sec:further-results}
\begin{figure}[h]
    \centering
    \begin{minipage}[c]{0.40\textwidth}
        \fbox{\includegraphics[width=\textwidth, trim={3cm 7.5cm 1.5cm 17.5cm}, clip]{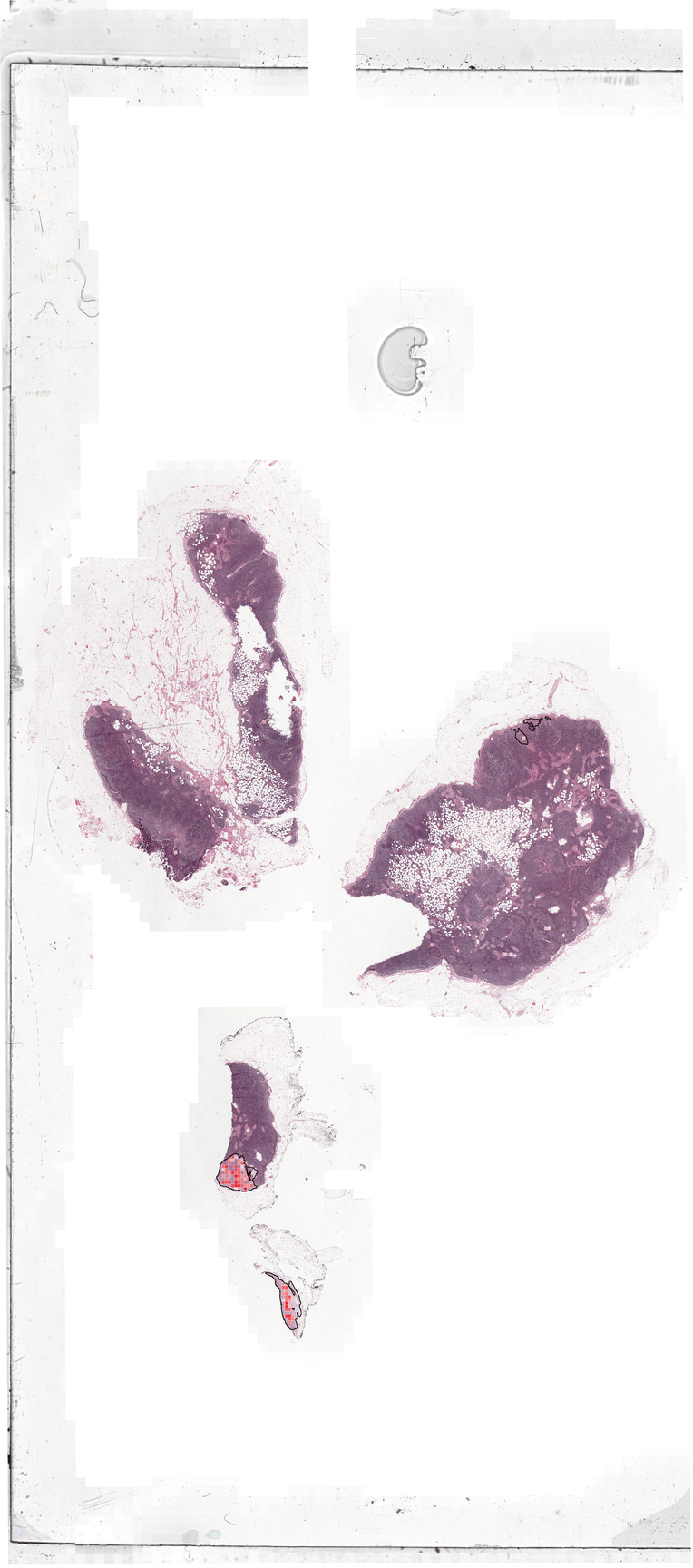}}
    \end{minipage}        
    \hfill
    \begin{minipage}[c]{0.40\textwidth}
        \fbox{\includegraphics[width=\textwidth]{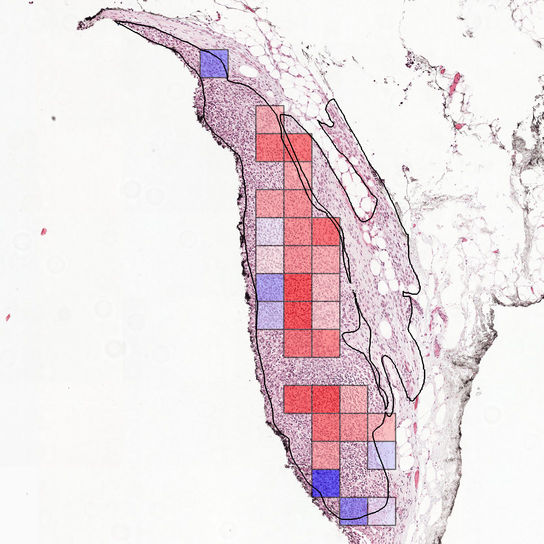}}\\
        \fbox{\includegraphics[width=\textwidth]{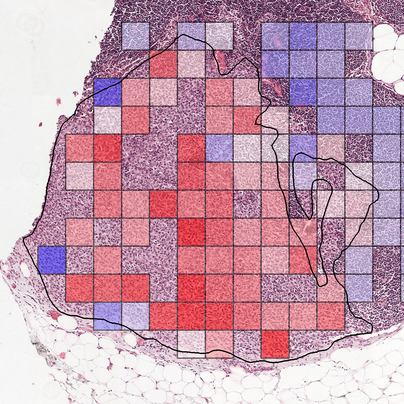}}\\
        \fbox{\includegraphics[width=\textwidth]{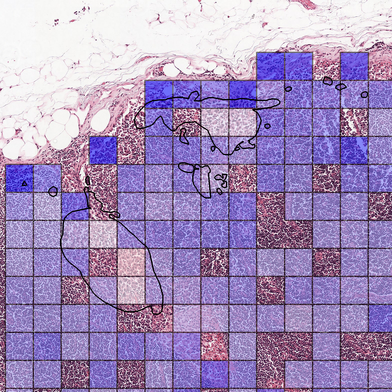}}
    \end{minipage}
    
    \caption{Visualization of metastasis detection on test image 2 of the \texttt{Camelyon-16} dataset 
            using our proposed approach. 
            \textbf{Left:} Full WSI at zoom level 6 with ground truth
            annotation of metastases shown via black border. Tiles with positive feature embeddings
            are colored from white to red according to their magnitude, with red 
            representing the largest magnitude.
            \textbf{Right:} Detail of metastases at zoom level 2 overlaid with classification 
            output of our proposed approach. Here, the output of all tested tiles are shown and
            colored according to their value, from blue to white to red, with blue representing
            the most negative values, and red the most positive.
            Tiles without color were not included when randomly selecting tiles for inference.}
\end{figure}

\begin{figure}[h]
    \centering
    \begin{minipage}[c]{0.45\textwidth}
        \fbox{\includegraphics[width=\textwidth]{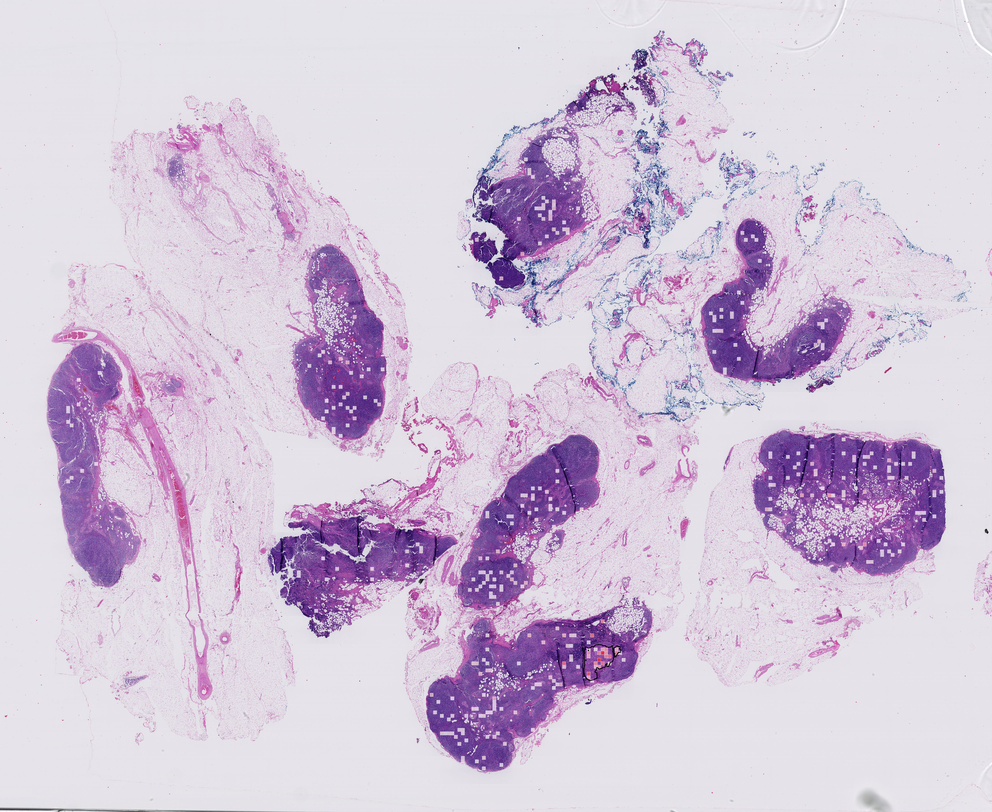}}
    \end{minipage}        
    \quad
    \begin{minipage}[c]{0.45\textwidth}
        \fbox{\includegraphics[width=\textwidth]{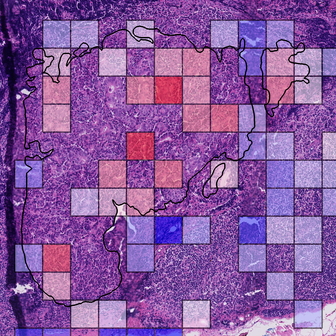}}\\
    \end{minipage}
    
    \caption{Visualization of metastasis detection on test image 92 of the \texttt{Camelyon-16} dataset 
            using our proposed approach. 
            \textbf{Left:} Full WSI at zoom level 6 with ground truth
            annotation of metastases shown via black border. Tiles with positive feature embeddings
            are colored from white to red according to their magnitude, with red 
            representing the largest magnitude.
            \textbf{Right:} Detail of metastases at zoom level 2 overlaid with classification 
            output of our proposed approach. Here, the output of all tested tiles are shown and
            colored according to their value, from blue to white to red, with blue representing
            the most negative values, and red the most positive.
            Tiles without color were not included when randomly selecting tiles for inference.}
\end{figure}

\end{document}